\documentclass{article}

\usepackage{microtype}
\usepackage{graphicx}
\usepackage{subfigure}
\usepackage{booktabs} % for professional tables

\usepackage{hyperref}

\usepackage[accepted]{icml2025}

\usepackage{amsmath}
\usepackage{amssymb}
\usepackage{mathtools}
\usepackage{amsthm}
\usepackage{colortbl} % import these two packages for colorful table
\usepackage{graphicx}
\usepackage{xcolor}         % colors
\usepackage{multirow}
\usepackage[utf8]{inputenc} % allow utf-8 input
\usepackage[T1]{fontenc}    % use 8-bit T1 fonts
\usepackage{hyperref}       % hyperlinks
\usepackage{url}            % simple URL typesetting
\usepackage{booktabs}       % professional-quality tables
\usepackage{amsfonts}       % blackboard math symbols
\usepackage{nicefrac}       % compact symbols for 1/2, etc.
\usepackage{enumitem}

\usepackage{wrapfig}
\usepackage{times}
\usepackage{latexsym}
\usepackage{booktabs}
\usepackage{makecell}
\usepackage{bm}
\usepackage{tabularray}
\usepackage{listings}
\usepackage{upquote}
\usepackage{graphicx}
\newcommand{\annotate}[3]{%
    #1\raisebox{-0.2ex}{\scriptsize\textcolor{#2}{#3}}%
}
\usepackage[most]{tcolorbox}
\newtcolorbox{promptbox}[1][]{
  colback=gray!10!white,
  colframe=gray!60!black,
  title={#1},
  fonttitle=\bfseries,
  boxrule=0.5mm,
  sharp corners,
  breakable,
  fontupper=\ttfamily\small, 
  #1
}
\newtcolorbox{modelcomparebox}[1][]{
  colback=gray!10!white,       % 背景色
  colframe=gray!60!black,      % 边框色
  title={#1},                  % 盒子标题（可选）
  fonttitle=\bfseries,         % 标题字体
  boxrule=0.5mm,               % 边框宽度
  sharp corners,               % 直角
  breakable,                   % 允许跨页
  fontupper=\ttfamily\small, 
  subtitle style={
    boxrule=0pt,               % 副标题无边框（只显示背景条）
    colback=gray!30!white,     % 分隔条背景色（比正文背景稍深）
    colupper=black,            % 分隔条文字颜色
    fontupper=\bfseries\sffamily\footnotesize, % 分隔条字体（无衬线加粗，易于区分）
    top=1mm, bottom=1mm,       % 分隔条上下的内边距
    sharp corners              % 分隔条直角
  }
}

\usepackage[capitalize, noabbrev]{cleveref}

\theoremstyle{plain}

\theoremstyle{definition}

\theoremstyle{remark}

\usepackage[textsize=tiny]{todonotes}

\icmltitlerunning{Self-Compression of Chain-of-Thought via Multi-Agent Reinforcement Learning}

\begin{document}
    \twocolumn[ \icmltitle{Self-Compression of Chain-of-Thought via Multi-Agent Reinforcement Learning}



    \icmlsetsymbol{equal}{*}
    \icmlsetsymbol{Corresponding author}{$\dagger$}

    \begin{icmlauthorlist}
        \icmlauthor{Yiqun Chen}{equal,ruc}
        \icmlauthor{Jinyuan Feng}{equal,casia}
        \icmlauthor{Wei Yang}{equal,usc}
        \icmlauthor{Meizhi Zhong}{xhs} 
        \icmlauthor{Zhengliang Shi}{sdu}
        \icmlauthor{Rui Li}{ruc} 
        \icmlauthor{Xiaochi Wei}{xhs}
        \icmlauthor{Yan Gao}{xhs}
        \icmlauthor{Yi Wu}{xhs}
        \icmlauthor{Yao Hu}{xhs}
        \icmlauthor{Zhiqiang Pu}{casia}
        \icmlauthor{Jiaxin Mao}{Corresponding author,ruc}
    \end{icmlauthorlist}

    \icmlaffiliation{ruc}{Renmin University of China}
    \icmlaffiliation{xhs}{Xiaohongshu Inc.}
    \icmlaffiliation{casia}{Institute of Automation,Chinese Academy of Sciences} 
    \icmlaffiliation{usc}{University of Southern California}
    \icmlaffiliation{sdu}{Shandong University}

    \icmlcorrespondingauthor{Jiaxin Mao}{maojiaxin@gmail.com}

    \icmlkeywords{Machine Learning, ICML}

    \vskip 0.3in ]



    \printAffiliationsAndNotice{\icmlEqualContribution} % otherwise use the standard text.

\begin{abstract}
The inference overhead induced by redundant reasoning undermines the interactive experience and severely bottlenecks the deployment of Large Reasoning Models. Existing reinforcement learning (RL)-based solutions tackle this problem by coupling a length penalty with outcome-based rewards. This simplistic reward weighting struggles to reconcile brevity with accuracy, as enforcing brevity may compromise critical reasoning logic. In this work, we address this limitation by proposing a multi-agent RL framework that selectively penalizes redundant chunks, while preserving essential reasoning logic. Our framework, Self-Compression via MARL (SCMA), instantiates redundancy detection and evaluation through two specialized agents: \textbf{a Segmentation Agent} for decomposing the reasoning process into logical chunks, and \textbf{a Scoring Agent} for quantifying the significance of each chunk. The Segmentation and Scoring agents collaboratively define an importance-weighted length penalty during training, incentivizing \textbf{a Reasoning Agent} to prioritize essential logic without introducing inference overhead during deployment. Empirical evaluations across model scales demonstrate that SCMA reduces response length by 11.1\% to 39.0\% while boosting accuracy by 4.33\% to 10.02\%. Furthermore, ablation studies and qualitative analysis validate that the synergistic optimization within the MARL framework fosters emergent behaviors, yielding more powerful LRMs compared to vanilla RL paradigms.
\end{abstract}

\section{Introduction}
\label{introduction}
Recent advancements in Large Reasoning Models (LRMs)~\cite{chen2025learning,guo2025deepseek} have demonstrated remarkable capabilities often characterized as 'thinking.' Specifically, LRMs employ detailed Chain-of-Thought (CoT) sequences to facilitate complex problem-solving through self-reflection, backtracking, and verification. The deep thinking capacities are predominantly elicited through Reinforcement Learning (RL)~\cite{shao2024deepseekmath,ramesh2024group}, but this paradigm simultaneously introduces significant efficiency bottlenecks. This issue arises because existing RL optimization is primarily driven by sparse, outcome-based binary rewards. Lacking fine-grained guidance on the intermediate reasoning process, models are prone to 'over-thinking'~\cite{chen2024not,cuadron2025danger}, where they generate prolonged paths cluttered with non-essential steps or repetitive verifications to maximize reward certainty. This redundancy imposes substantial inference latency for deployment, necessitating methods that guide models to 'think less' by synthesizing concise, robust, and accurate reasoning chains.

Current methods, which we discuss extensively in Appendix~\ref{appendix:related_works}, primarily address these limitations by incorporating length-based penalties to curb reasoning redundancy~\cite{cheng2025optimizing,tu2025learning,dai2025s,zeng2025done,hou2025thinkprune}. However, such solutions suffer from misaligned credit assignment: they indiscriminately penalize the total sequence length rather than intrinsic redundancy of intermediate reasoning steps~\cite{zelikman2022star}. This undifferentiated strategy poses a significant risk: models may inadvertently sacrifice critical reasoning steps essential for success just to satisfy length constraints, thereby compromising final accuracy. Therefore, a pivotal challenge remains: how to distinguish and preserve high-value reasoning steps at a fine-grained level, while precisely eliminating redundancy and meaningless repetition. The key to overcoming this bottleneck lies in two aspects: first, decomposing the continuous reasoning process into independent logical chunks; and second, accurately quantifying the substantive contribution of each chunk to the final derivation. Regrettably, existing single-agent RL paradigms struggle to achieve this, as they inherently lack the mechanisms for fine-grained structural modeling and decoupled value estimation. This necessitates a paradigm shift towards Multi-Agent Reinforcement Learning (MARL), where distinct agents can collaborate to achieve simultaneous optimization of reasoning generation and fine-grained process control.
% Additionally, we provide a section for related work in Appendix A to introduce these methods more specifically.

To bridge this gap, we propose Self-Compression via MARL (SCMA), an end-to-end multi-agent framework designed to achieve fine-grained compression of the thinking process. SCMA innovatively reformulates the compression task as a collaborative game, coordinating three specialized roles: the Reasoning Agent explore the solution space to generate reasoning paths; the Segmentation Agent structurally parses the reasoning paths into discrete logical chunks; and the Scoring Agent quantifies the substantive contribution of each chunk to the final derivation. In this setup, standard length penalties are replaced by an importance-weighted length penalty derived collaboratively by the Segmentation and Scoring agents. This feedback incentivizes the Reasoning Agent to selectively discard redundancy while preserving essential logic. Crucially, since the Reasoning Agent’s ability to balance correctness with conciseness hinges entirely on the fidelity of this dynamic feedback, the three roles are intrinsically coupled. Consequently, SCMA employs a shared reward to drive the co-evolution of all agents by multi-agent group relative policy optimization (GRPO), ensuring their objectives remain strictly aligned towards generating high-quality, efficient reasoning chains.

In summary, our contributions can be concluded as follows:
\begin{itemize}[leftmargin=1em, itemsep=1pt, topsep=1pt, parsep=0pt]
\item \textbf{Framework Proposal:} We propose Self-Compression via MARL (SCMA), a framework that reformulates the Chain-of-Thought compression task by transitioning from the conventional single-agent RL to a MARL training paradigm, enabling fine-grained compression with no test-time overhead.
\item \textbf{Mechanism Design:} We design a collaborative optimization mechanism governed by a multi-agent GRPO objective, incorporating an importance-weighted length penalty as a unified reward signal to drive the co-evolution of all agents.
\item \textbf{Experimental Verification:} We empirically demonstrate that SCMA reduces reasoning length by \textbf{11.1\%--39.0\%} while simultaneously improving accuracy by \textbf{4.33\%--10.02\%}. Ablation studies further validate the superiority of MARL joint optimization, which significantly outperforms decoupled baselines relying on single-agent architectures.
\end{itemize}

% We design a collaborative mechanism comprising Reasoning, Segmentation, and Scoring Agents that are jointly optimized via MARL with a shared reward signal, facilitating the simultaneous enhancement of reasoning conciseness, segmentation rationality, and scoring accuracy.
%%%%%%%%%%% 示例
\iffalse
\item We propose SCMA, a MARL-based self-compression framework with a Reasoning Agent for CoT generation, a Segmentation Agent for semantic chunking, and a Scoring Agent for per-chunk importance, enabling fine-grained compression with no test-time overhead.

\item We develop a shared-payoff multi-agent GRPO objective with an importance-weighted compression reward that penalizes low-importance chunks under a correctness gate, plus format-based stabilization, enabling adaptive length control without sacrificing essential reasoning.
\fi
%%%%%%%%%%%

\begin{figure*}[t]
\centering
\includegraphics[width=0.9\textwidth]{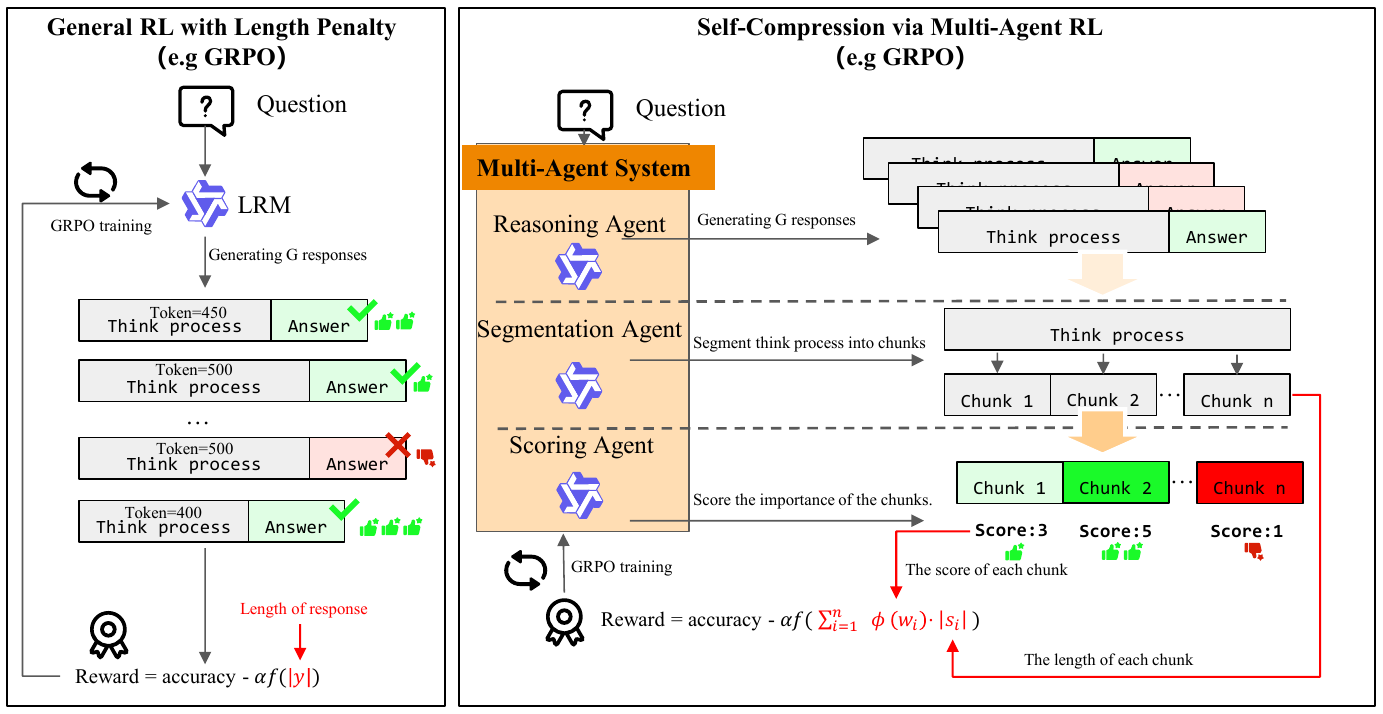}
\caption{Overview of SCMA Compared to general RL with length penalty. \textbf{(Left)} The general RL calculates rewards by penalizing the length of the thinking process directly. \textbf{(Right)} The SCMA employs an importance-weighted length penalty within a multi-agent system. }
\label{fig:framework}
\vspace{-0.2cm}
\end{figure*}
\section{Preliminary}
\subsection{General RL with Length Penalty}
\label{subsec:general_RL}
In post-training based on RL, LLM is modeled as a policy $\pi_\theta$ that autoregressively generates a response $y$ given a prompt $x$. To compress the reasoning process, existing solutions typically incorporate a length penalty, formulating the reward function as $R(y|x) = R_{\text{acc}}(y|x) - \alpha f(|y|)$. $R_{\text{acc}}$ denotes outcome-based rewards for correctness, $\alpha$ is the penalty strength coefficient, and $f(\cdot)$ applies group-wise normalization to the response length $|y|$, typically taking a linear or exponential form~\cite{cheng2025optimizing,zhang2025grpo}. Consequently, the RL objective is to optimize $\theta$ for maximizing the expected reward:
\begin{equation}
    \mathcal{J}(\theta) = \mathbb{E}_{x \sim \mathcal{D}, y \sim \pi_\theta(\cdot|x)} \left[ R_{\text{acc}}(y|x) - \alpha f(|y|) \right]
    \label{eq:single RL}
\end{equation}

\subsection{Group Relative Policy Optimization (GRPO)}
\label{sec:GRPO}
GRPO~\cite{shao2024deepseekmath} streamlines the alignment
of large language models by eliminating the critic network inherent to PPO~\cite{schulman2017proximal}, thereby
reducing memory overhead and avoiding value approximation instability. Specifically, GRPO estimates the advantages by sampling a group of responses $\{y_{i}\}_{i=1}^{G}$ and normalizing the rewards within group. The advantage for each token $t$ in the $i$-th
response is computed as
$\hat{A}_{i,t}= \frac{r_{i} - \text{mean}(\mathbf{r})}{\text{std}(\mathbf{r})}$,
where $\mathbf{r}$ is the reward vector for the group. The GRPO objective incorporates a clipping mechanism and a KL-divergence penalty
to stabilize policy updates and prevent excessive deviation from the reference policy $\pi_{ref}$:
\begin{equation}
    \begin{aligned}
        \mathcal{J}_{\text{GRPO}}(\theta) & = \mathbb{E}_{x \sim P(x), \{y_i\}_{i=1}^G \sim \pi_{\theta_{\text{old}}}(\cdot|x)}\frac{1}{G}\sum_{i=1}^{G} \frac{1}{|y_i|}\sum_{t=1}^{|y_i|}   \\
                                          & \Bigg\{ \min \Bigl[ r_{i,t}(\theta) \hat{A}_{i,t},\ \operatorname{clip}\bigl( r_{i,t}(\theta), 1-\epsilon, 1+\epsilon \bigr) \hat{A}_{i,t}\Bigr] \\
                                          & \quad - \beta D_{\text{KL}}\bigl[ \pi_{\theta} \parallel \pi_{\text{ref}}\bigr] \Bigg\},
    \end{aligned}
    \label{eq:GRPO}
\end{equation}

\noindent
where the importance sampling ratio is defined as:
\begin{equation}
    r_{i,t}(\theta) = \frac{\pi_{\theta}(y_{i,t}\mid x, y_{i,<t})}{\pi_{\theta_{\text{old}}}(y_{i,t}\mid
    x, y_{i,<t})}.
\end{equation}
Based on $\{y_{i}\}_{i=1}^{G}$ sampled from the
behavior policy $\pi_{\theta_{\text{old}}}$, the optimization is constrained by the clipping hyperparameter
$\epsilon$ and the KL-divergence penalty coefficient $\beta$. 
\section{Method}
\subsection{An Overview of SCMA}
As illustrated in Fig~\ref{fig:framework}, SCMA departs from the scalar penalties of general RL by instantiating three functionally distinct agents within a multi-agent system to achieve fine-grained compression. Crucially, all agents share parameters $\theta$ from a single base LLM $\pi_{\text{base}}$ to achieve internal reasoning compression, a process we term \textbf{Self-Compression}. Under this paradigm, the multi-agent framework serves as a training-only auxiliary, enabling the exclusive deployment of the Reasoning Agent during inference without incurring additional computational overhead. Within the Self-Compression process, the agents are assigned the following specialized roles:
\begin{enumerate}
    \item The \textbf{Reasoning Agent} ($\pi_\theta^{\text{reason}}$), which explores the solution space to generate an initial reasoning path $y$: 
    \begin{equation}
        y \sim \pi_\theta^{\text{reason}}(\cdot \mid x)
        \label{eq:reason}
    \end{equation}
    \item The \textbf{Segmentation Agent} ($\pi_\theta^{\text{seg}}$) structurally parses the reasoning path $y$ into $n$ discrete logical chunks:
    \begin{equation}
        \{s_1, \dots, s_n\}\sim\pi_\theta^{\text{seg}}(y)
        \label{eq:seg}
    \end{equation}
    \item The \textbf{Scoring Agent} ($\pi_\theta^{\text{score}}$) then evaluates the inferential significance of each chunk, assigning an importance score $w_i$ that quantifies its essentiality in deriving the final correct solution:
    \begin{equation}
       \{w_1, \dots, w_n\}\sim \pi_\theta^{\text{score}}(\{s_1, \dots, s_n\})
       \label{eq:score}
    \end{equation}
\end{enumerate}
Building upon the multi-agent system, the Segmentation and Scoring agents collaboratively define an \textbf{importance-weighted length penalty} to achieve fine-grained compression. This distinguishes SCMA from the general RL approach in Section~\ref{subsec:general_RL} and mitigates the inherent risk of sacrificing critical reasoning steps, which is a common failure mode in monolithic sequence-level optimization. Specifically, the reformulated total reward is defined as:
\begin{equation}
    R(y|x) = R_{\text{acc}}(y|x) - \alpha f\left(\sum_{i=1}^{n} \textcolor{red}{\underbrace{\phi(w_i) \cdot |s_i|}_{\text{importance-weighted length}}}\right),
\end{equation}
where $|s_i|$ denotes the length of the $i$-th segment, and $\phi(w_i)$ is a weighting function that modulates the penalty scale based on the segment's importance score. Intuitively, $\phi(\cdot)$ is designed as a monotonically decreasing function to inversely map importance to penalty weight: a redundant segment ($w_i \to 0$) is assigned a high weight (e.g., $\phi(w_i) \approx 5$) to enforce compression, whereas a critical segment ($w_i \to 5$) receives a negligible weight (e.g., $\phi \approx 0$) to exempt it from the length penalty. This mechanism encourages the agent to preserve pivotal reasoning logic when compression.

We frame the optimization of SCMA as a Multi-Agent Reinforcement Learning (MARL) problem, where the reformulated total reward serves as a shared global payoff for all participating agents.  This collective incentive ensures that the core objective of producing accurate and concise responses is inherently synchronized with the individual functional goals of each agent. In this process, the Reasoning Agent learns to generate accurate solutions while eliminating redundant reasoning steps. Simultaneously, the Segmentation Agent evolves to perform more rational and fine-grained partitioning, enabling the Scoring Agent to effectively assess the importance of the thinking process. We formalize the joint trajectory as a structured tuple that captures the sequential dependencies among the three agents:
\begin{equation}
\tau = \left\{
\begin{aligned}
    & \text{"response"} : y, \\
    & \text{"chunks"} : \mathcal{S} = \{s_1, s_2, \dots, s_n\}, \\
    & \text{"scores"} : \mathcal{W} = \{w_1, w_2, \dots, w_n\}
\end{aligned}
\right\}
\label{eq:trajectory}
\end{equation}
The transition $(x, \tau, R(y|x))$ drives the MARL optimization, with each agent \{$\pi_\theta^{\text{reason}}$, $\pi_\theta^{\text{seg}}$, $\pi_\theta^{\text{score}}$\} deriving its own advantage $\hat{A}_k(\tau, x)$ from the shared global reward.

\subsection{Detailed Configuration for MARL}
Grounded in the theory of Markov Games~\cite{littman1994markov}, the SCMA framework is formulated as a tuple:
\begin{equation}
    \mathcal{M} = \langle \mathcal{G}, \mathcal{V}, \mathcal{O}, \mathcal{A}, \mathcal{R} \rangle,
\end{equation}
where $\mathcal{G} = \{\pi_\theta^{\text{reason}}, \pi_\theta^{\text{seg}}, \pi_\theta^{\text{score}}\}$ represents the set of specialized agents responsible for reasoning, segmentation, and scoring, respectively.
$V$ represents a vocabulary of the underlying large language model, serving as the foundational set for both $\mathcal{O}$ and $\mathcal{A}$.
$\mathcal{O} = \{ \mathcal{O}_i \}_{i \in \mathcal{G}}$ represents the collection of each local observations, $\mathcal{A} = \prod_{i \in \mathcal{G}} \mathcal{A}_i$ is the joint action space, and $\mathcal{R}$ is a reward function shared for all agents. Generally, both the observation space and the action space of each agent are defined over the shared vocabulary $\mathcal{V}$ of the language model, treating the generation process as a sequential token-level decision problem~\cite{ouyang2022training}. In practice,  the effective observation space and action space of each agent are constrained by its role-specific prompt~\cite{li2023camel}. The concrete instantiations of the observation space, action space, and reward function for different agents are specified in the following.

\textbf{The observation space of each agent:} As formalized in Eq.~\ref{eq:reason},~\ref{eq:seg} and ~\ref{eq:score}, the observation spaces exhibit a sequential dependency chain, where the output action of one agent constitutes the primary context for the next. Each agent's observation $o_i \in \mathcal{O}_i$ is constructed by concatenating its role-specific prompt $P_i$ with the relevant task context. Specifically, the \textbf{Reasoning Agent} integrates the role-specific prompt $P_{\text{reason}}$ with the question $q$ as $x$ to generate a reasoning path $y$. Subsequently, the \textbf{Segmentation Agent} treats reasoning path $y$ and its role-specific prompt $P_{\text{seg}}$ as its observation, parsing it into structured logical chunks. Finally, the \textbf{Scoring Agent} observes the sequence of discrete logical chunks $\{s_1, \dots, s_n\}$ and role-specific prompt $P_{\text{score}}$ to evaluate the importance of each chunk. $\mathcal{O}$ is given by:
\begin{equation}
\begin{aligned}
\mathcal{O}^{\text{reason}} &= \{x, P_{\text{reason}}\}  \\
\mathcal{O}^{\text{seg}} &= \{y,P_{\text{seg}} \mid y \sim \pi_\theta^{\text{reason}}(\cdot \mid x)\} \\
\mathcal{O}^{\text{score}} &= \{S,P_{\text{score}} \mid S \sim \pi_\theta^{\text{seg}}(y)\}
\end{aligned}
\end{equation}
The concrete implementation details and full templates for $P_{\text{reason}}$, $P_{\text{seg}}$, and $P_{\text{score}}$ are provided in Appendix~\ref{app:prompts}.

\textbf{The action space of each agent:} Corresponding to the sequential observation process, the action spaces of the agents are defined by their specific generative objectives and formatting constraints. The \textbf{Reasoning Agent} operates within an unconstrained solution search space, generating a natural language sequence $y$ from the Kleene closure~\cite{kleene1956representation} of the model vocabulary $\mathcal{V}^*$. The \textbf{Segmentation Agent} performs a structural transformation, partitioning $y$ into a sequence of steps delimited by \texttt{<seg>} tags, subject to the constraint that the composition of these segments reconstructs the original path. Finally, the \textbf{Scoring Agent} maps each segment to a discrete scalar value $w_i \in \{1, \dots, 5\}$, encapsulated by \texttt{<score>} tags.
\begin{equation}
\begin{aligned}
\mathcal{A}^{\text{reason}} &= \{y \mid y \in \mathcal{V}^*\} \\
\mathcal{A}^{\text{seg}} &= \left\{ \{\texttt{<seg>} s_i \texttt{</seg>}\}_{i=1}^n \;\middle|\; \text{s.t.} \bigoplus_{i=1}^n s_i = y \right\} \\
\mathcal{A}^{\text{score}} &= \left\{\{ \texttt{<score>} w_i \texttt{</score>} \right\}_{i=1}^n\mid w_i \in \{1,..., 5\} \}
\end{aligned}
\label{eq:action}
\end{equation}
where $\mathcal{V}^*$ represents the set of all possible token sequences generated by the language model. The constraint in $\mathcal{A}^{\text{seg}}$ ensures lossless parsing, where the union of the segmented contents $s_i$ must equate to the original reasoning path $y$.

\textbf{The shared reward function:} We introduce an importance-weighted length penalty to balance reasoning accuracy with conciseness, the design philosophy of which is formulated in Eq.~\ref{eq:reward}. Specifically, the weighting function $\phi(w_i) = 5 - w_i$ maps the importance score $w_i \in \{1, \dots, 5\}$ to a penalty coefficient, ensuring that critical segments with higher scores incur a reduced penalty. The function $f(\cdot)$ then applies group-wise normalization to the weighted length by computing $\sum \phi(w_i) \cdot |s_i|$ and normalizing it by the maximum total length among all correct reasoning paths in the candidate set $\mathcal{C}$. The total reward $R(y|x)$ is calculated as:
\begin{equation}
\begin{cases} 
R_{\text{acc}}(y|x)+\alpha(1 - \frac{\sum_{i=1}^{n}(5-w_i)\cdot|s_i|}{\max_{j \in \mathcal{C}}\left(\sum_{m=1}^{n_j}|s_{j,m}|\right)}), & \text{if } y \in \mathcal{C} \\
0, & \text{if } y \notin \mathcal{C}.
\end{cases}
\label{eq:reward}
\end{equation}
To mitigate bias from intrinsic problem complexity, the penalty is normalized against the maximum weighted length within the training batch $\mathcal{C}$. This relative scaling ensures equitable regularization across different tasks, while $\alpha$ balances the penalty’s influence on the total reward. We show in Sec.~\ref{theoretical_analysis} that this objective is equivalent to maximizing the expected $R_{\text{acc}}$ under a weighted length constraint.

In addition to the shared reward, we further define agent-specific format rewards ($R_{\text{fmt}}$) to ensure structural consistency and stabilize multi-agent training. A binary reward is assigned for strict adherence to the following XML-style tag protocols and the specific constraints of the action space: the Reasoning Agent must enclose thinking process within \texttt{<think>} tags; the Segmentation Agent must comprehensively demarcate segments using \texttt{<seg>} tags; and the Scoring Agent is required to assign scores to every segment within \texttt{<score>} tags.

\subsection{Multi-Agent GRPO Optimization}
As discussed in Section~\ref{sec:GRPO}, the standard GRPO is inherently designed for single-agent settings, primarily focusing on calculating advantages for the reasoning agent. In this work, we extend GRPO to MARL context to collaboratively optimize the policies of three agents: $\mathcal{G} = \{\pi_\theta^{\text{reason}}, \pi_\theta^{\text{seg}}, \pi_\theta^{\text{score}}\}$. Adopting a configuration similar to Multi-Agent PPO (MAPPO) in \textit{StarCraft II}~\cite{yu2022surprising,feng2025ma2rl}, our Multi-Agent GRPO leverages a shared global reward and employs a parameter-sharing strategy to enhance sample efficiency and foster synergistic collaboration among agents. Consequently, all three agents are instantiated from a single base LLM $\pi_{\text{base}}$ and are distinguished only by role-specific prompts $P_{\text{reason}}$, $P_{\text{seg}}$, $P_{\text{score}}$. The pseudocode for the multi-agent GRPO optimization is presented in Algorithm~\ref{alg:scma}. The procedure initiates with the Multi-Agent Rollout phase, where the input data propagates sequentially through the Reasoning, Segmentation, and Scoring agents to generate cooperative reasoning chains. Following the trajectory collection, the Multi-Agent Optimization phase is executed. In this stage, group relative advantages are computed for each agent, and the shared parameters $\theta$ are updated by maximizing the GRPO objective for each agent respectively.
\begin{algorithm}[!h]
\caption{SCMA: Multi-Agent Training with GRPO}
\label{alg:scma}
\textbf{Input:} Initial shared parameters $\theta$ (for $\pi^{\text{reason}}, \pi^{\text{seg}}, \pi^{\text{score}}$), task prompts $\mathcal{D}$, group size $G$ \\
\textbf{Output:} $\mathcal{G} = \{\pi_\theta^{\text{reason}}, \pi_\theta^{\text{seg}}, \pi_\theta^{\text{score}}\}$
\begin{algorithmic}[1]
    \FOR{step = $1, \dots, M$}
        \STATE Sample a batch $\mathcal{D}_b$ from $\mathcal{D}$
        \STATE \textbf{// Phase 1: Multi-Agent Rollout}
        \FOR{each question $x \in \mathcal{D}_b$}
            \STATE Sample $G$ cooperative chains for group $\{k\}_{k=1}^G$:
            \STATE \quad 1. \textbf{Reasoning:} $y_k \sim \pi_{\theta}^{\text{reason}}(\cdot \mid x)$ (Eq. \ref{eq:reason})
            \STATE \quad 2. \textbf{Segmentation:} $S_k \sim \pi_{\theta}^{\text{seg}}(\cdot \mid y_k)$ (Eq. \ref{eq:seg})
            \STATE \quad 3. \textbf{Scoring:} $W_k \sim
            \pi_{\theta}^{\text{score}}(\cdot \mid S_k)$ (Eq.~\ref{eq:score})
            \STATE \textbf{// Phase 2: Global Reward \& Group Advantage}
            \STATE Compute global reward $R(y|x)$ using Eq.~\ref{eq:reward} and the format reward $R_{\text{fmt}}$
            \STATE Store transition $(x, \tau, R(y|x))$ (Eq.~\ref{eq:trajectory})
            \STATE Compute group advantages $\hat{A}_{i,t}^{\text{reason}}, \hat{A}_{i,t}^{\text{seg}}, \hat{A}_{i,t}^{\text{score}}$ for each agent $\{\pi_\theta^{\text{reason}}, \pi_\theta^{\text{seg}}, \pi_\theta^{\text{score}}\}$ 
        \ENDFOR
        \STATE \textbf{// Phase 3: Multi-Agent Optimization}
        \FOR{$\hat{A}_{i,t}^{\text{reason}}, \hat{A}_{i,t}^{\text{seg}}, \hat{A}_{i,t}^{\text{score}}$}
            \STATE Update shared parameters $\theta$ by maximizing the GRPO objective (Eq.~\ref{eq:GRPO}):
        \ENDFOR
    \ENDFOR
    \STATE \textbf{return} $\mathcal{G} = \{\pi_\theta^{\text{reason}}, \pi_\theta^{\text{seg}}, \pi_\theta^{\text{score}}\}$
\label{SCMA-code}
\end{algorithmic}
\end{algorithm}
\begin{table*}[t]
\centering
\scriptsize
% \small
\caption{Performance and reasoning length of SCMA and baselines across multiple large reasoning models.}
\label{table1}
\setlength\tabcolsep{4.5pt} 
\renewcommand{\arraystretch}{1}
\begin{tabular}{@{}lccccccccccll@{}} 
\toprule
 \multirow{2}{*}{\textbf{Method}} 
 & \multicolumn{2}{c}{\textbf{GSM8K}} & \multicolumn{2}{c}{\textbf{MATH500}} & \multicolumn{2}{c}{\textbf{AIME24}} & \multicolumn{2}{c}{\textbf{AIME25}} & \multicolumn{2}{c}{\textbf{AMC23}}    & \multicolumn{2}{|c}{\textbf{Overall}} \\
    & {Acc} & {Tokens} & Acc & Tokens & {Acc}  & {Tokens} & {Acc} & {Tokens} & {Acc} & {Tokens}  & Acc & Tokens \\ 
\hline
\multicolumn{13}{l}{{\cellcolor[rgb]{0.957,0.957,0.957}}\textit{\textbf{DeepSeek-R1-Distill-Qwen-1.5B}}} \\
\textit{Vanilla} & 85.97 & 1496 & 78.40 & 3801 & 33.30 & 9237 & 20.00 & 9478 & 62.50 & 6261  & \multicolumn{1}{|l}{{56.03}} & {6054}  \\
\textit{GRPO} & {87.71} & {1063} & {79.00} & {3773} & {40.00} & {8224} & {30.00} & {9273} & {70.00} & {6069}  & \multicolumn{1}{|l}{{\textbf{\annotate{61.34}{red}{\;\,+5.31\%}}}} & {\textbf{\annotate{5680}{blue}{\;\,-6.2\%}}}  \\
\textit{LC-R1\_LP} & {79.68} & {451} & {77.40} & {3092} & {36.60} & {8656} & {26.60} & {8504} & {67.5} & {5414}  & \multicolumn{1}{|l}{{\textbf{\annotate{57.55}{red}{\;\,+1.52\%}}}} & {\textbf{\annotate{5223}{blue}{\;\,-13.7\%}}}  \\
\textit{RL+LP} & {81.50} & {556} & {76.20} & {2718} & {30.00} & {7683} & {26.60} & {7851} & {62.5} & {4671}  & \multicolumn{1}{|l}{{\textbf{\annotate{55.36}{blue}{\;\,-0.67\%}}}} & {\textbf{\annotate{4695}{blue}{\;\,-22.4\%}}}  \\
\textbf{SCMA (Ours)} & {86.20} & {661} & {79.00} & {3551} & {36.66} & {8672} & {30.00} & {8518} & {70.00} & {5504}  & \multicolumn{1}{|l}{{\textbf{\annotate{60.36}{red}{\;\,+4.33\%}}}} & {\textbf{\annotate{5381}{blue}{\;\,-11.1\%}}}  \\
\hline

\multicolumn{13}{l}{{\cellcolor[rgb]{0.957,0.957,0.957}}\textit{\textbf{DeepSeek-R1-Distill-Qwen-7B}}} \\
\textit{Vanilla} & 92.40 & 1833 & 86.40 & 3080 & 50.00 & 8726 & 36.66 & 8195 & 82.50 & 5229  & \multicolumn{1}{|l}{{69.59}} & {5412}  \\
\textit{GRPO} & {93.17} & {941} & {86.80} & {3094} & {46.66} & {7987} & {40.00} & {8626} & {85.00} & {4981}  & \multicolumn{1}{|l}{{\textbf{\annotate{70.32}{red}{\;\,+0.73\%}}}} & {\textbf{\annotate{5125}{blue}{\;\,-5.2\%}}}  \\
\textit{LC-R1\_LP} & {82.10} & {121} & {86.40} & {2696} & {50.00} & {7951} & {40.00} & {7970} & {90.00} & {4631}  & \multicolumn{1}{|l}{{\textbf{\annotate{69.70}{red}{\;\,+0.11\%}}}} & {\textbf{\annotate{4673}{blue}{\;\,-13.6\%}}}  \\
\textit{RL+LP} & {91.28} & {376} & {82.80} & {2416} & {46.66} & {7781} & {33.33} & {7586} & {90.00} & {4561}  & \multicolumn{1}{|l}{{\textbf{\annotate{68.81}{blue}{\;\,-0.77\%}}}} & {\textbf{\annotate{4544}{blue}{\;\,-16.0\%}}}  \\
\textbf{SCMA (Ours)} & {93.02} & {588} & {86.20} & {2654} & {60.00} & {7473} & {43.33} & {7657} & {90.00} & {4521}  & \multicolumn{1}{|l}{{\textbf{\annotate{74.51}{red}{\;\,+4.92\%}}}} & {\textbf{\annotate{4578}{blue}{\;\,-15.4\%}}}  \\
\hline

\multicolumn{13}{l}{{\cellcolor[rgb]{0.957,0.957,0.957}}\textit{\textbf{Qwen3-4B}}} \\
\textit{Vanilla} & {94.49} & 1320 & {86.60} & 4300 & {43.33} & 9805 & {30.00} & 9958 & {82.50} & 6943  & \multicolumn{1}{|l}{{67.38}} & {6465}  \\
\textit{GRPO} & {94.92} & 1001 & {87.20} & 3911 & {60.00} & 8680 & {40.00} & 9356 & {90.00} & 6441  & \multicolumn{1}{|l}{{\textbf{\annotate{74.42}{red}{\;\,+7.04\%}}}} & {\textbf{\annotate{5877}{blue}{\;\,-9.0\%}}}  \\
\textit{LC-R1\_LP} & {93.70} & {378} & {88.2} & {2892} & {56.66} & {8096} & {36.66} & {7819} & {87.5} & {4032}  & \multicolumn{1}{|l}{{\textbf{\annotate{72.54}{red}{\;\,+5.16\%}}}} & {\textbf{\annotate{4643}{blue}{\;\,-28.1\%}}}  \\
\textit{RL+LP} & {94.01} & {381} & {87.2} & {2249} & {53.33} & {7295} & {43.33} & {8040} & {87.5} & {4217}  & \multicolumn{1}{|l}{{\textbf{\annotate{73.07}{red}{\;\,+5.69\%}}}} & {\textbf{\annotate{4436}{blue}{\;\,-31.3\%}}}  \\
\textbf{SCMA (Ours)} & {94.16} & 351 & {88.00} & 1629 & {60.00} & 7100 & {43.33} & 7402 & {95.00} & 3242  & \multicolumn{1}{|l}{{\textbf{\annotate{76.09}{red}{\;\,+8.70\%}}}} & {\textbf{\annotate{3944}{blue}{\;\,-39.0\%}}}  \\
\hline
 
\multicolumn{13}{l}{{\cellcolor[rgb]{0.957,0.957,0.957}}\textit{\textbf{Qwen3-8B}}} \\

\textit{Vanilla} & 95.40 & 1888 & 85.80 & 4512 & 40.00 & 8944 & 33.33 & 9501 & 72.50 & 6928   & \multicolumn{1}{|l}{{65.40}} & 6354 \\
\textit{GRPO} & 95.98 & 1342 & 88.40 & 4182 & 56.67 & 8789 & 46.66 & 9344 & 90.00 & 6844  & \multicolumn{1}{|l}{{\textbf{\annotate{75.54}{red}{\;\,+10.14\%}}}} & {\textbf{\annotate{6100}{blue}{\;\,-3.99\%}}}  \\
\textit{LC-R1\_LP} & {95.98} & {447} & {88.80} & {2841} & {56.6} & {8304} & {40.00} & {7636} & {82.50} & {4937}  & \multicolumn{1}{|l}{{\textbf{\annotate{72.77}{red}{\;\,+7.37\%}}}} & {\textbf{\annotate{4833}{blue}{\;\,-23.9\%}}}  \\
\textit{RL+LP} & {95.75} & {432} & {88.4} & {2539} & {53.33} & {7973} & {46.67} & {8425} & {90.00} & {3878}  & \multicolumn{1}{|l}{{\textbf{\annotate{74.83}{red}{\;\,+9.43\%}}}} & {\textbf{\annotate{4649}{blue}{\;\,-26.8\%}}}  \\
\textbf{SCMA (Ours)} & 94.99 & 369 & 89.20 & 1999 & 60.00 & 6475 & 43.33 & 7004 & 89.60 & 3599  & \multicolumn{1}{|l}{{\textbf{\annotate{75.42}{red}{\;\,+10.02\%}}}} & {\textbf{\annotate{3889}{blue}{\;\,-38.8\%}}}  \\
\end{tabular}
\label{tab:main_result}
\vspace{-0.4cm}
\end{table*}
\section{Experiment}
We design our experiments to answer the following questions: \textbf{RQ1:} Can SCMA outperform existing baselines, particularly length-penalized RL methods, by achieving effective reasoning compression without compromising solution accuracy? \textbf{RQ2:} How does the hyperparameter $\alpha$ influence the training effectiveness and the stability of SCMA? \textbf{RQ3:} Is the multi-agent cooperative optimization essential for our framework? \textbf{RQ4:} How does the fine-grained compression capability emerge during training?

\subsection{Experiment Setting}
\textbf{Dataset, Baselines and LLM Models}: We evaluate our framework across a spectrum of difficulty levels, from fundamental arithmetic to competition mathematics, using GSM8K~\cite{cobbe2021training}, MATH500~\cite{hendrycks2021measuring}, AMC23~\cite{li2024numinamath}, and AIME24/25~\cite{li2024numinamath}. We benchmark our approach against GRPO and two established baselines that employ length penalties: LC-R1\_LP~\cite{cheng2025optimizing} and RL+LP~\cite{arora2025training}. Detailed specifications for the datasets and baselines are elaborated in Appendix~\ref{app:baselines}. We select DeepSeek-R1-Distill-Qwen (1.5B/7B)~\cite{deepseekai2025deepseekr1incentivizingreasoningcapability} and Qwen3 (4B/8B)~\cite{qwen3technicalreport} series as base models due to widespread adoption. Our analysis focuses on accuracy (Acc), and response length (Tokens).

% baselines:
\textbf{Implementation Details}: We implement SCMA based on the open-source \texttt{verl} framework~\cite{sheng2024hybridflow}, utilizing GRPO as the underlying optimization backbone for SCMA. To ensure experimental consistency, all models are fine-tuned using the GSM8K dataset, which contains 7,473 training samples. Regarding hyperparameter configurations, we set the training context size to 11K, the batch size to 256, the $\alpha$ to 0.1, and the learning rate to $1 \times 10^{-6}$. For GRPO-specific settings, we configure the number of rollouts to 5 and the KL penalty coefficient to 0.001. During validation, we adopt a sampling decoding strategy with a temperature of 0.6 and a top-$p$ value of 0.95. Answer extraction and verification are enabled by default in verl.
\subsection{Main results}
\textbf{Answer of Question 1:}
We evaluate SCMA against multiple baselines across five benchmark datasets: GSM8K, MATH500, AIME24, AIME25, and AMC23. As illustrated in Table~\ref{tab:main_result}, SCMA consistently achieves superior overall accuracy across models of varying parameter scales. Compared to the vanilla reasoning model, SCMA demonstrates remarkable advantages in both efficiency and performance, achieving an average reduction in thinking process ranging from 11.1\% to 39.0\% while simultaneously boosting accuracy by 4.33\% to 10.02\%.

\noindent\textbf{Advantages on Length and Accuracy:} Experimental results demonstrate that SCMA's efficacy amplifies with model scale, expanding from an 11.1\% length reduction on DeepSeek-1.5B to nearly 39\% on Qwen3-8B (decreasing from 6,354 to 3,889 tokens), where the MATH500 task notably exhibits a reduction exceeding 50\% (from 4,512 to 1,999 tokens) alongside an accuracy boost to 89.20\%. This scalability, which allows larger models to achieve high compression rates while maintaining high accuracy, confirms SCMA's capacity to excise redundant reasoning chains and unleash the latent potential of high-performance models.

\noindent\textbf{Cost-Performance Trade-Off:} Generally, length penalties-based RL methods often struggle to balance efficiency and accuracy. For instance, while RL+LP reduces length by 16.0\% on DeepSeek-7B, it causes accuracy to drop below the baseline (68.81\% $<$ 69.59\%). In contrast, SCMA achieves a superior trade-off: on Qwen3-4B, SCMA not only reduces length by 39.0\% (significantly outperforming LC-R1\_LP) but also attains the highest accuracy of 76.09\%. This demonstrates that SCMA optimizes reasoning paths rather than merely truncating them, significantly enhancing logical density while eliminating redundancy.

\noindent\textbf{Out-of-Distribution Generalization} Although trained exclusively on GSM8K, SCMA exhibits exceptional generalization capabilities across the remaining benchmarks. It not only significantly improves accuracy on unseen challenging problems (e.g., using DeepSeek-7B achieves 60.00\% on AIME24, surpassing GRPO's 46.66\%) but also maintains high efficiency across tasks. For example, in the MATH500 evaluation using Qwen3-4B, the response length plummets from 4,300 to 1,629 tokens (a reduction of over 62\%). This confirms that SCMA does not overfit the training data; rather, it has genuinely mastered a generalizable and compact reasoning paradigm.

\noindent\textbf{Inference Efficiency:} SCMA functions strictly as a training-time auxiliary via shared parameters $\theta$. Auxiliary agents are deactivated during deployment, leaving the Reasoning Agent to operate autonomously. This guarantees zero additional computational overhead, while the induced brevity further reduces total decoding latency.
\begin{figure}[t]
    \centering
    \subfigure[GSM8K]{\includegraphics[width=0.49\linewidth]{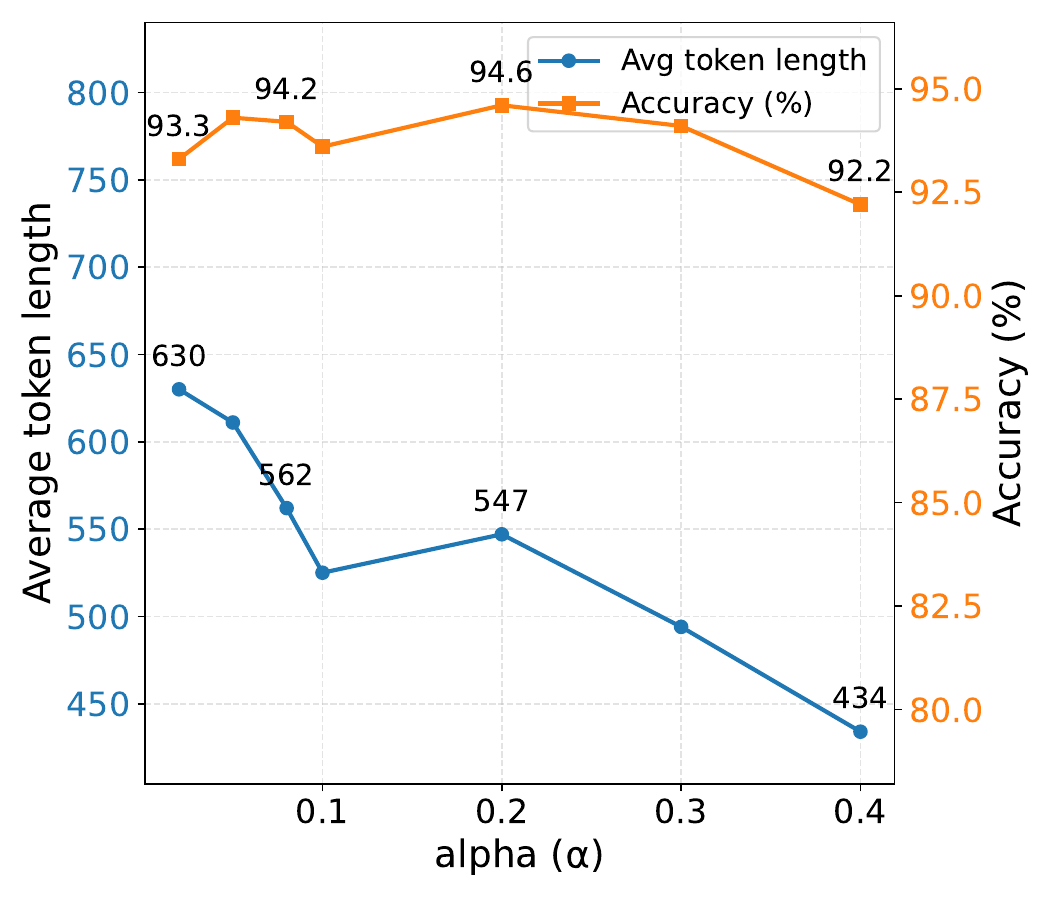}}
    \subfigure[MATH500]{\includegraphics[width=0.49\linewidth]{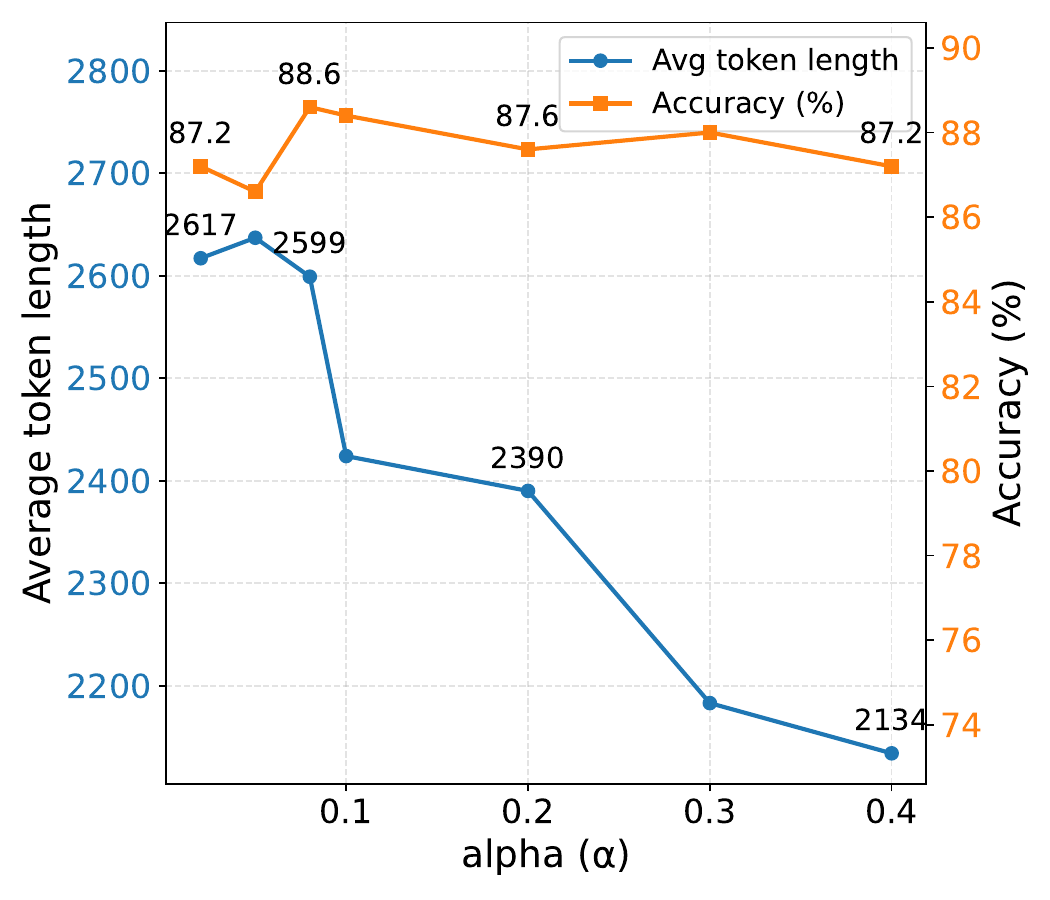}}
    \caption{Average accuracy and response length of Qwen3-4B trained with different $\alpha$}
    \label{hyperparameter}
    \vspace{-0.4cm}
\end{figure}
\begin{figure}[t]
    \centering
    \includegraphics[width=0.85\linewidth]{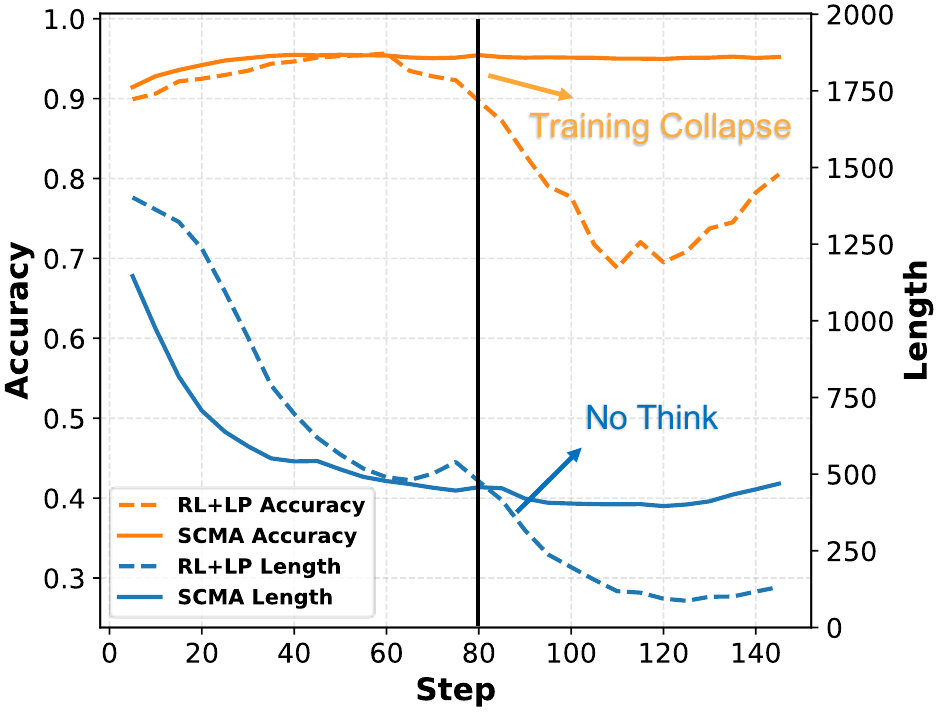}
    \caption{Training curves of SCMA and RL+LP. During training, the RL+LP model suffers from training collapse, with response length dropping significantly, indicating a "No Think" pattern.}
    \label{fig:training_dynamic}
    \vspace{-0.4cm}
\end{figure}
\begin{figure*}[ht]
\centering
\includegraphics[width=0.8\textwidth]{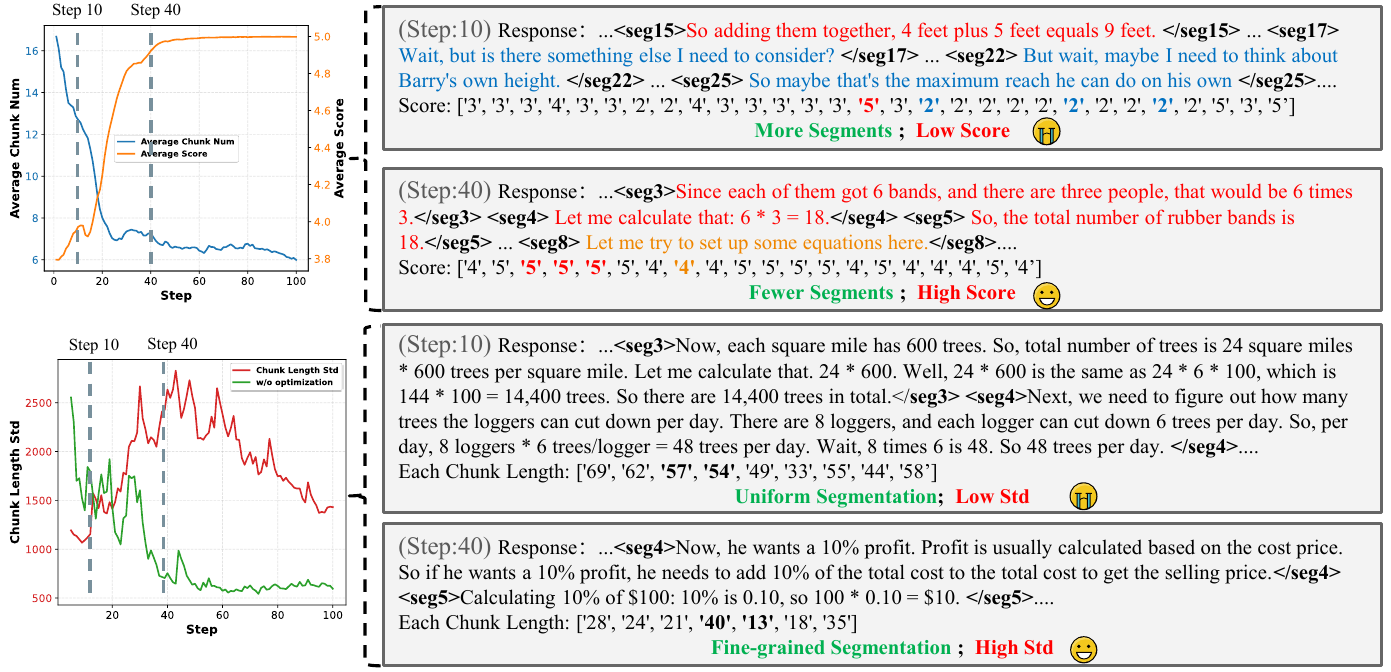}
\caption{Evolution of the Scoring and Segmentation Agents. \textbf{(Left)} Curves showing the Average Score (orange), Average Chunk Num (blue), Chunk Length Std (red), the Chunk Length Std if Segmentation Agents are not optimized, for shot w/o optimization (green) over training steps. \textbf{(Right)} Case of agents at Step 10 versus Step 40, including segmentation points and score annotations. }
\label{fig:metric_case}
\vspace{-0.2cm}
\end{figure*}
\begin{table*}[ht]
\centering
\scriptsize
% \small
\caption{Ablation study on component optimization. Here, only the reasoning agent (Qwen-4B) is optimized, while the Segmentation and Scoring agents remain frozen and are implemented using off-the-shelf Qwen models (8B, 4B, and 1.7B) via API. }
\label{table1}
\setlength\tabcolsep{4.5pt} 
\renewcommand{\arraystretch}{1}
\begin{tabular}{@{}lccccccccccll@{}} 
\toprule
 \multirow{2}{*}{\textbf{Method}} 
 & \multicolumn{2}{c}{\textbf{GSM8K}} & \multicolumn{2}{c}{\textbf{MATH500}} & \multicolumn{2}{c}{\textbf{AIME24}} & \multicolumn{2}{c}{\textbf{AIME25}} & \multicolumn{2}{c}{\textbf{AMC23}}    & \multicolumn{2}{|c}{\textbf{Overall}} \\
    & {Acc} & {Tokens} & Acc & Tokens & {Acc}  & {Tokens} & {Acc} & {Tokens} & {Acc} & {Tokens}  & Acc & Tokens \\ 
\hline
\textbf{SCMA (Ours)} & {94.54} & {588} & {88.20} & {3360} & {50.00} & {8495} & {46.66} & {9016} & {87.50} & {5704}  & \multicolumn{1}{|l}{73.38} & {5432}  \\
\multicolumn{13}{l}{{\cellcolor[rgb]{0.957,0.957,0.957}}\textit{\textbf{Ablation Studies}}} \\
\textit{w/o\_optimization\textbf{-gpt4o}} & 93.40 & 702 & 87.20 & 3541 & 60.00 & 8418 & 43.33 & 9568 & 80.00 & 5830  & \multicolumn{1}{|l}{{\textbf{\annotate{72.78}{blue}{-0.59}}}} & {\textbf{\annotate{5611}{red}{+3.31\%}}}  \\
\textit{w/o\_optimization\textbf{-8b}} & 92.83 & 798 & 86.6 & 3636 & 53.33 & 8989 & 43.33 & 9216 & 90.00 & 6176  & \multicolumn{1}{|l}{{\textbf{\annotate{73.21}{blue}{-0.17}}}} & {\textbf{\annotate{5763}{red}{+6.09\%}}}  \\
\textit{w/o\_optimization\textbf{-4b}} & 94.76 & 855 & 87.80 & 3675 & 53.33 & 8802 & 36.66 & 9335 & 82.50 & 5752  & \multicolumn{1}{|l}{{\textbf{\annotate{71.01}{blue}{-2.37}}}} & {\textbf{\annotate{5683}{red}{+4.62\%}}}  \\
\textit{w/o\_optimization\textbf{-1.7b}} & 94.76 & 984 & 85.80 & 4044 & 50.00 & 8873 & 36.66 & 10007 & 85.00 & 6497 & \multicolumn{1}{|l}{{\textbf{\annotate{70.44}{blue}{-2.94}}}} & {\textbf{\annotate{6081}{red}{+11.94\%}}}  \\
\hline
\end{tabular}
\label{tab:ablation_result}
\vspace{-0.4cm}
\end{table*}
\subsection{Hyperparameter Study and Training Stability}
\textbf{Answer of Question 2:} Fig.~\ref{hyperparameter} presents the sensitivity analysis of SCMA regarding the length penalty hyperparameter $\alpha$. The results indicate that as $\alpha$ increases from 0.05 to 0.40, the inference length decreases significantly across both datasets, exemplified by a reduction from 630 to 434 tokens on GSM8K, which underscores the high efficiency of SCMA in compressing the thinking process. Crucially, the model accuracy remains highly stable with only minor fluctuations despite the substantial shortening of the reasoning chains. This demonstrates that SCMA successfully retains core reasoning information while eliminating redundancy, thereby validating the robust performance of the framework under varying compression intensities.

\noindent\textbf{Analysis of Stability} To investigate the training dynamics and stability of SCMA compared to length penalties-based RL methods, we visualize the training curves of both methods in Fig.~\ref{fig:training_dynamic}. As observed, RL+LP is prone to training collapse during the mid-to-late stages, where the agent aggressively compresses response length to maximize rewards once accuracy converges, inadvertently driving the model into a "No Think" pattern. In contrast, our SCMA achieves a superior trade-off between accuracy and conciseness by employing a scoring agent that selectively penalizes only redundant reasoning while strictly preserving core logic. Appendix~\ref{app:stability} details the length penalty and length of SCMA to justify stability and the efficacy of dynamic penalty.

\subsection{Ablation Studies}
\textbf{Answer of Question 3:} To verify the necessity of the multi-agent RL optimization within the SCMA framework, we conducted ablation studies by freezing the Segmentation and Scoring agents. Specifically, we replaced the optimized internal modules with closed-source GPT-4o and off-the-shelf Qwen models (1.7B, 4B, and 8B) invoked via API, while keeping the Reasoning agent (Qwen-4B) optimized via GRPO. The results are presented in Table~\ref{tab:ablation_result}.

\textbf{Superiority of Cooperative Optimization.} 
As shown in Table~\ref{tab:ablation_result}, our proposed SCMA achieves the highest overall accuracy of \textbf{73.38\%}, significantly outperforming the unoptimized variants. When the Segmentation and Scoring agents are replaced with a frozen model of the same size (\textit{w/o\_optimization-4b}), the overall accuracy drops by \textbf{2.37\%} to 71.01\%, and token consumption increases by \textbf{4.62\%}. This demonstrates that joint parameter optimization aligns the agents' objectives more effectively than simply pipeline-chaining pre-trained models.

\textbf{Beating Larger Models with Efficiency.} 
Notably, SCMA even surpasses the variant utilizing a larger 8B model and GPT-4o model as the auxiliary agent (\textit{w/o\_optimization-8b}, \textit{w/o\_optimization-gpt4o}). For example, despite the 8B model's stronger intrinsic capabilities, the lack of cooperative tuning results in a lower overall accuracy (\textbf{73.21\%}) and a significantly higher computational cost (\textbf{+6.09\%} tokens). This indicates that the domain-specific collaboration established through our multi-agent optimization is more critical than the raw scale of the auxiliary models.

\textbf{Impact of Segmentation and Scoring Quality.} 
The performance gap widens further when smaller models are used. The \textit{w/o\_optimization-1.7b} variant exhibits the lowest performance (70.44\% accuracy) and the highest inefficiency (\textbf{+11.94\%} token usage). This decline underscores that the quality of segmentation and scoring directly bounds the reasoning capabilities. Without rational chunk segmentation and accurate importance scoring provided by the segmentation agent and the scoring agent, the Reasoning agent struggles to maintain high-quality reasoning chains.

\subsection{Evolutionary Dynamics of SCMA}
\textbf{Answer of Question 4:} By monitoring the evolution of key metrics and analyzing qualitative samples, as illustrated in Fig.~\ref{fig:metric_case}, we observe a significant emergence of fine-grained compression capability as training steps increase:

\noindent\textbf{Semantic Concentration \& Redundancy Pruning} As depicted in Fig.~\ref{fig:metric_case} (Top Left \& Top Right), a distinct evolutionary pattern emerges between Step 0 and Step 100: the Average Chunk Num (blue line) declines sharply, while the Average Score (orange line) ascends synchronously towards convergence. This negative correlation implies a synergistic co-optimization between the Reasoning Agent ($\pi^{\text{reason}}_\theta$) and the Scoring Agent ($\pi^{\text{score}}_\theta$). Qualitatively, the model transitions away from the low-scoring, hesitant, and repetitive fragments typical of early training (e.g., the redundant \textit{``...Wait, but is there something else...''} observed at Step 10), evolving instead to produce highly condensed reasoning kernels. We interpret this phenomenon through an information-theoretic lens as the model learning to maximize information entropy per unit fragment. By employing \textbf{Semantic Compression}, the model prunes redundancy to converge the reasoning process into critical steps with elevated \textbf{Semantic Density} and substantial logical contribution, simultaneously optimizing reasoning efficiency and quality.

\noindent\textbf{Content-Adaptive Fine-grained Segmentation} As illustrated in Fig.~\ref{fig:metric_case} (Bottom Left \& Bottom Right), the trajectory of the Chunk Length Std (red line) offers critical insights into the evolution of the Segmentation Agent. Initially maintaining a low level, the standard deviation undergoes a sharp ascent after Step 20. This statistical trend signifies a behavioral shift in the Segmentation Agent ($\pi^{\text{seg}}_\theta$): \textit{a transition from ''uniform segmentation'' to ''fine-grained segmentation''.} Specifically, the low standard deviation observed in the early phase (e.g., Step 10) implies that the model tends towards segment text into equal-length chunks while remaining agnostic to the underlying semantic importance. Conversely, the high variance observed at Step 40 indicates that the model has acquired the capability to dynamically allocate chunk length based on the cognitive load of the reasoning content. As corroborated by the qualitative samples, the agent learns to allocate extended text blocks to encapsulate complete logical flows (such as complex numerical calculations or derivations), while assigning concise segments to simple transitions or definitions. This behavior demonstrates that the model has developed a sophisticated capacity for \textbf{Content-Adaptive Fine-grained Segmentation}, regulating information flow based on semantic density rather than executing uniform textual segmentation.

\noindent\textbf{Qualitative Evolution Summary} A comparative analysis of the textual outputs between Step 10 and Step 40 (Fig.~\ref{fig:metric_case}, Right) delineates the clear evolutionary trajectory of the model's compression capabilities: \textbf{(Step 10)} The reasoning path is notably verbose, saturated with low-confidence (Low Score) exploratory phrases. The segmentation granularity appears rigid and indiscriminate, lacking semantic focus. \textbf{(Step 40)} The total number of chunks ($n$) decreases, while the importance score ($w_i$) assigned to each segment ($s_i$) sees a substantial elevation. This shift indicates that the model has learned to precisely identify logical breakpoints, achieving efficient compression while rigorously preserving the structural integrity of the reasoning chain.

\section{Conclusion}
This work addresses the issue of redundant reasoning chains in Large Reasoning Models. Distinct from existing single-agent Reinforcement Learning methods that rely on coarse-grained length penalties, we construct a Multi-Agent Reinforcement Learning framework utilizing fine-grained reward shaping. Empirical evaluations confirm that our framework significantly shortens thinking process while maintaining reasoning performance. Furthermore, qualitative analysis delineates the clear evolutionary trajectory of the model's compression capabilities, validating the synergy inherent in the multi-agent system. Future work will scale this paradigm towards large-scale, heterogeneous systems, thereby empowering the resolution of increasingly complex settings.
    \nocite{langley00}

    \bibliography{example_paper}
    \bibliographystyle{icml2025}

    %%%%%%%%%%%%%%%%%%%%%%%%%%%%%%%%%%%%%%%%%%%%%%%%%%%%%%%%%%%%%%%%%%%%%%%%%%%%%%%
    %%%%%%%%%%%%%%%%%%%%%%%%%%%%%%%%%%%%%%%%%%%%%%%%%%%%%%%%%%%%%%%%%%%%%%%%%%%%%%%
    % APPENDIX
    %%%%%%%%%%%%%%%%%%%%%%%%%%%%%%%%%%%%%%%%%%%%%%%%%%%%%%%%%%%%%%%%%%%%%%%%%%%%%%%
    %%%%%%%%%%%%%%%%%%%%%%%%%%%%%%%%%%%%%%%%%%%%%%%%%%%%%%%%%%%%%%%%%%%%%%%%%%%%%%%
    \newpage
    \appendix
    \onecolumn
    \section{Related Work}
\label{appendix:related_works}
\noindent\textbf{Efficient Reasoning in LRMs} Large Reasoning Models (LRMs) often depend on lengthy chains of thought (CoT) to achieve strong performance. However, such exhaustive reasoning can be unnecessary for low- to medium-complexity tasks, leading to overthinking and substantial computational overhead~\cite{chen2024not}. To address this issue, recent studies propose early-exit mechanisms, adaptive inference, and dynamic prompting.

Early-exit methods aim to terminate generation once sufficient confidence is reached, leveraging signals such as internal state transitions or uncertainty thresholds~\cite{liao2025fractured,team2024qwen2,qiao2025concise}, or by assessing the consistency between intermediate reasoning steps and candidate answers~\cite{liu2025answer}. Other works adopt implicit stopping strategies without explicit termination criteria~\cite{dai2025s}. While effective and often training-free, these approaches rely heavily on heuristic rules and exhibit limited generalization. In contrast, adaptive inference methods dynamically adjust reasoning depth or patterns to better match task complexity. Representative strategies include reward-guided control of reasoning length~\cite{jiang2025think,luo2025adar1} and dynamic switching between reasoning modes~\cite{zhang2025othink,zhang2025adaptthink,tu2025learning}. Similarly, dynamic prompting reduces redundant reasoning through inference-time prompt engineering~\cite{han2025token,muennighoff2025s1,xu2025chain,ma2025reasoning}. Nevertheless, these approaches typically depend on predefined priors over task difficulty or output length, making it challenging to learn a unified and intrinsic mechanism for regulating inference behavior.

\noindent\textbf{RL-based Post-Training for Efficient Reasoning} With Reinforcement Learning with Verifiable Rewards (RLVR) emerging as a dominant paradigm for post-training LRMs~\cite{yang2025maestro,chang2025survey,yang2025learning,yang2025toward}, it has been observed that optimizing solely for final answer correctness often induces excessively long chains of thought, exacerbating overthinking and increasing inference cost~\cite{yue2025don,zhu2025towards}. Consequently, recent studies have explored explicitly incorporating efficiency considerations into the RL reward function to mitigate redundant reasoning while preserving accuracy.

One line of work introduces length-based penalties or constraints to directly account for reasoning cost. For instance, ~\cite{cheng2025optimizing,arora2025training,aggarwal2025l1,cui2025stepwise} penalize the output length during RL training, encouraging the generation of more concise reasoning traces and balancing accuracy with computational efficiency. O1-Pruner~\cite{luo2025o1}, ThinkPrune~\cite{hou2025thinkprune} and Kimi~\cite{team2025kimi} further propose pruning-oriented RL strategies, applying stronger negative rewards to long or inefficient trajectories to guide the model toward early termination or compression of reasoning steps when appropriate. Although these methods have demonstrated empirical success, they typically treat efficiency coarsely as total sequence length and operate mainly over parallel chains, offering limited control over stepwise reasoning within a single chain. This can result in reduced accuracy when compressing reasoning chains or suppressing redundant steps, highlighting the need for RL approaches that more finely penalize unnecessary reasoning while preserving correctness.
\section{Prompt Designs for SCMA}
\label{app:prompts}

In this section, we provide the detailed prompt templates used for the three agents defined in our framework: the Reasoning Agent ($P_{\text{reason}}$), the Segmentation Agent ($P_{\text{seg}}$), and the Scoring Agent ($P_{\text{score}}$).

\subsection{Reasoning Agent Prompt ($P_{\text{reason}}$)}
The Reasoning Agent generates the initial Chain-of-Thought solution path $y$ given the input problem $x$. We utilize a standard zero-shot Chain-of-Thought prompt to encourage step-by-step reasoning.

\begin{promptbox}[title=Reasoning Prompt ($P_{\text{reason}}$)]
Let's think step by step and output the final answer within \textbackslash boxed\{\}.
\end{promptbox}

\subsection{Segmentation Agent Prompt ($P_{\text{seg}}$)}
The Segmentation Agent takes the generated reasoning path $y$ and segments it into minimal reasoning units. The prompt focuses on preserving the semantic completeness of each unit without omitting any original text.

\begin{promptbox}[title=Segmentation Prompt ($P_{\text{seg}}$)]
You will be given a passage of text. Please do the following:

1. Segment the text into ``minimal reasoning units.'': A ``minimal reasoning unit'' is the smallest chunk that is logically or semantically complete enough to be understood or evaluated on its own (it can be a sentence, a clause, or a short phrase, but do not break the text so much that you destroy natural semantics).

Your segmentation must cover the entire original text, without omissions and without changing the order.
\end{promptbox}

\subsection{Scoring Agent Prompt ($P_{\text{score}}$)}
The Scoring Agent evaluates the importance of each segmented unit derived from the previous step. It assigns a score on a scale of 1 to 5 and formats the final output with specific tags.

\begin{promptbox}[title=Scoring Prompt ($P_{\text{score}}$)]
For each segmented unit, assign an importance score based on how important that unit is for the overall reasoning process, using a 1–5 scale:
1: Almost not important; has negligible impact on the overall reasoning
2: Contributes something, but below average importance
3: Average level of contribution
4: Above-average level of contribution
5: Very important; plays a key or central role in the overall reasoning

Output format requirements:

You must fully reproduce the original text in your output, but divided into the segmented units, each wrapped with the following tags:
For the 1st unit:
<seg1>content of the first segment</seg1><score1>score</score1>
For the 2nd unit:
<seg2>content of the second segment</seg2><score2>score</score2>
And so on, until the nth unit:
<segn>content of the n-th segment</segn><scoren>score</scoren>

The concatenation of all <seg*> contents (in order) must reconstruct the original text exactly in terms of wording and order. Minor differences in whitespace or line breaks are acceptable, but you must not add, delete, or paraphrase any words.

Output only the tagged structure described above. Do not include any explanations, comments, or extra text.

Here is the text to process: [Input Text]
\end{promptbox}

\section{Theoretical Analysis: Formulation as Constrained Optimization}
\label{theoretical_analysis}
Instead of viewing the reward function merely as a heuristic length penalty, we formulate our task as a \textbf{Constrained Markov Decision Process (CMDP)}. Our objective is to maximize the accuracy $R_{\text{acc}}$ subject to a constraint on the ``weighted length cost,'' where the cost is inversely related to the importance score $w_i$.

\subsection{Primal Problem Definition}
Let $C(y)$ denote the weighted length cost of a generated sequence $y$. We define this cost based on the premise that tokens with higher importance ($w_i$) should incur a lower penalty. The optimization problem is defined as:

\begin{equation}
\begin{aligned}
\max_{\theta} \quad & \mathcal{J}(\theta) = \mathbb{E}_{y \sim \pi_\theta(\cdot|x)} \left[ R_{\text{acc}}(y|x) \right] \\
\text{s.t.} \quad & \mathbb{E}_{y \sim \pi_\theta(\cdot|x)} \left[ C(y) \right] \leq \beta
\end{aligned}
\label{eq:cmdp}
\end{equation}
where $\beta$ is a pre-defined budget for redundancy. Consistent with our reward design, the weighted cost $C(y)$ is explicitly defined as:
\begin{equation}
C(y) = \frac{\sum_{i=1}^{n}(5-w_i)\cdot|s_i|}{L_{\text{norm}}}
\label{eq:cost_function}
\end{equation}
Here, $L_{\text{norm}} = \max_{j \in \mathcal{C}}\left(\sum_{m=1}^{n_j}|s_{j,m}|\right)$ acts as a normalization factor to scale the cost.

\subsection{Lagrangian Relaxation}
To solve this constrained problem, we employ the method of Lagrange multipliers. We introduce a Lagrange multiplier $\lambda \geq 0$ to incorporate the constraint into the objective function, converting the primal problem into the following unconstrained dual problem:

\begin{equation}
\min_{\lambda \geq 0} \max_{\theta} \mathcal{L}(\theta, \lambda) = \mathbb{E}_{y \sim \pi_\theta} \left[ R_{\text{acc}}(y|x) - \lambda \cdot (C(y) - \beta) \right]
\end{equation}

By fixing $\lambda$ as a hyperparameter and rearranging the terms, the equivalent maximization objective for the policy $\pi_\theta$ becomes:
\begin{equation}
\max_{\theta} \mathbb{E}_{y \sim \pi_\theta} \left[ R_{\text{acc}}(y|x) - \lambda \cdot C(y) + \lambda \beta \right]
\label{eq:lagrangian_obj}
\end{equation}

\subsection{Connection to Proposed Reward}
We now demonstrate that our proposed reward function in Eq.~\ref{eq:reward} is a specific instantiation of the Lagrangian objective derived above. Recall our reward function design (for $y \in \mathcal{C}$):
\begin{equation}
R(y|x) = R_{\text{acc}}(y|x) + \alpha \left( 1 - \underbrace{\frac{\sum_{i=1}^{n}(5-w_i)\cdot|s_i|}{L_{\text{norm}}}}_{C(y)} \right)
\end{equation}
Expanding this equation yields:
\begin{equation}
R(y|x) = \underbrace{R_{\text{acc}}(y|x)}_{\text{Accuracy}} - \underbrace{\alpha}_{\lambda} \cdot \underbrace{C(y)}_{\text{Weighted Cost}} + \underbrace{\alpha}_{\text{Constant Bias } (\lambda\beta)}
\label{eq:expanded_reward}
\end{equation}

By comparing Eq.~\ref{eq:expanded_reward} with the Lagrangian objective in Eq.~\ref{eq:lagrangian_obj}, we establish the following theoretical equivalences:
\begin{itemize}
    \item The hyperparameter $\alpha$ functions as the \textbf{Lagrange multiplier} $\lambda$, controlling the trade-off between accuracy and the information density constraint.
    \item The term $(5-w_i)$ in $C(y)$ implements the weighted constraint, where preserving high-importance information ($w_i \uparrow$) incurs a lower cost ($C(y) \downarrow$).
    \item The constant term $+\alpha$ corresponds to $\lambda \beta$, serving as a reward baseline that reduces variance during training without altering the optimal policy direction.
\end{itemize}

\section{Evaluation datasets and baselines}
\label{app:baselines}
To evaluate the mathematical reasoning capabilities of our framework across a wide spectrum of difficulty levels that range from foundational arithmetic to elite competition mathematics, we utilize five representative benchmarks:
\begin{itemize}
    \item \textbf{GSM8K} consists of several thousand grade-school math word problems that require multi-step reasoning and the application of basic arithmetic operations.
    \item \textbf{MATH500} serves as a curated subset of the MATH dataset, encompassing high-school level challenges across diverse domains such as algebra, geometry, and probability.
    \item \textbf{AMC23} features problems from the 2023 American Mathematics Competitions, which demand both solid subject knowledge and a high degree of logical flexibility.
    \item \textbf{AIME24} is derived from the 2024 American Invitational Mathematics Examination and represents a substantial leap in difficulty to test the model's precision in handling long-chain mathematical deductions.
    \item \textbf{AIME25} includes the most recent 2025 AIME problems, serving as a frontier benchmark to assess the performance limits on highly sophisticated and cognitively demanding mathematical tasks.
\end{itemize}

To evaluate the effectiveness of our approach, we compare it against several RL-based baselines that focus on reasoning efficiency and reward optimization:
\begin{itemize}
    \item \textbf{GRPO}~\cite{shao2024deepseekmath}: Group Relative Policy Optimization is a reinforcement learning framework that eliminates the necessity of a critic model by estimating advantages through reward normalization across a set of responses generated from the same prompt.
    \item \textbf{LC-R1\_LP}: This baseline stems from the LC-R1~\cite{cheng2025optimizing} framework, which originally integrates a Length Reward for conciseness and a Compress Reward for removing invalid thinking processes. In our experimental setup, we adapt this framework by excluding the LC-Extractor and exclusively employing its length penalty function.
    \item \textbf{RL+LP}~\cite{arora2025training}: This configuration implements a length penalty mechanism that assigns reward values based on the deviation of each correct response's length from the group mean. This approach specifically penalizes excessive verbosity by reducing the rewards for correct but unnecessarily long responses, thereby encouraging the model to generate more efficient reasoning chains.
\end{itemize}

\section{Analysis of SCMA Stability and Dynamic Penalty Efficacy}
\label{app:stability}
\begin{figure}[h]
    \centering
    \includegraphics[width=0.45\linewidth]{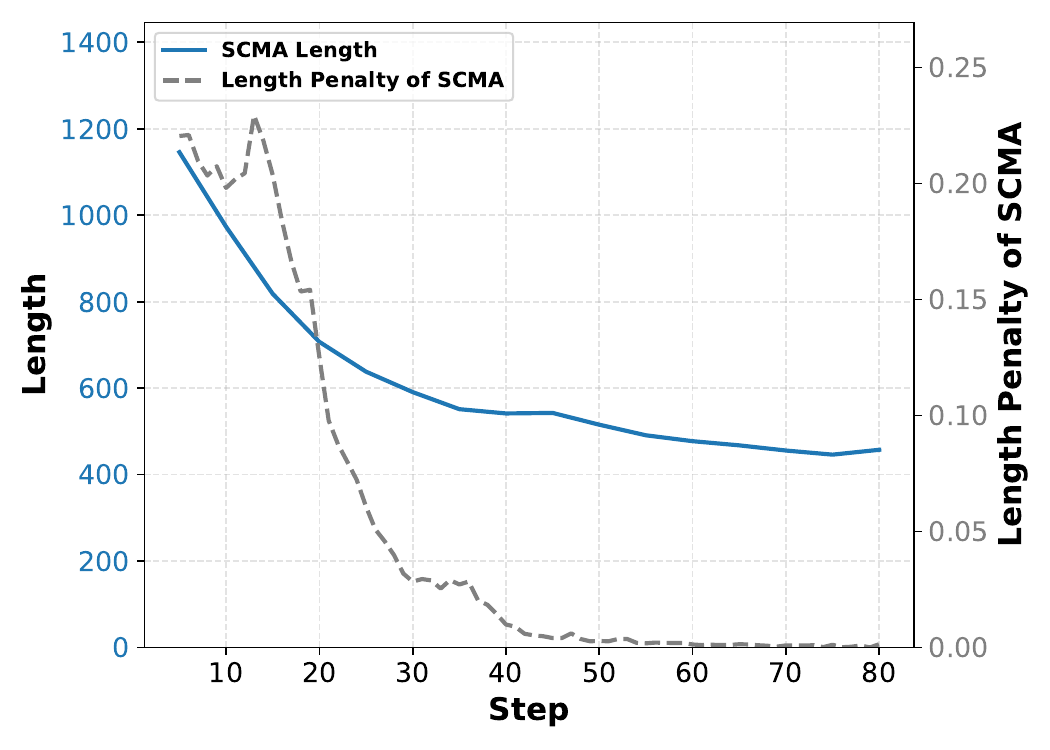}
    \caption{Dynamic evolution of response length and length penalty during SCMA training. The solid blue line (left axis) denotes the mean response length, while the dashed grey line (right axis) represents the importance-weighted length penalty. The figure illustrates how the dynamic penalty modulates response length over training steps to ensure convergence and stability.}
    \label{fig:stability}
\end{figure}
As illustrated in Fig.~\ref{fig:stability}, the importance-weighted length penalty demonstrates an autonomous dynamic adjustment process that significantly influences model convergence. The training process can be analyzed in two distinct phases to elucidate the underlying mechanism of this self-adaptive regulation:
\begin{itemize}
    \item \textbf{The initial phase}, spanning from step 0 to step 30, is characterized by a high-penalty regime that drives rapid length compression. During this period, the model frequently generates responses containing redundant or low-quality semantic chunks. As indicated by the grey dashed line, the length penalty remains at an elevated level above 0.20 and reaches a local maximum around step 15. This behavior occurs because the importance-weighted length penalty identifies that these low-scoring chunks contribute minimally to the overall logic of the response, thereby necessitating a substantial penalty. Consequently, under this intense negative feedback, the model effectively eliminates invalid content, resulting in a sharp decrease in the SCMA length from approximately 1150 to 600 as shown by the solid blue line.
    \item \textbf{The subsequent phase}, occurring between step 30 and step 80, transitions into a state where core logical structures are retained under a diminished penalty. As redundant information is successfully filtered, the model begins to focus on the production of high-quality outputs. The grey dashed line exhibits a precipitous decline toward zero during this interval. This reduction is attributed to the fact that the remaining segments are primarily high-scoring chunks essential to the core reasoning of the response. The importance-weighted length penalty identifies these components as necessary and ceases to apply significant pressure for further compression. As the penalty approaches zero, the SCMA length stabilizes into a plateau between 450 and 500, marking the end of the rapid reduction phase.
\end{itemize}
The results confirm that the importance-weighted length penalty functions as a sophisticated adaptive regulator rather than a static hyperparameter. By imposing rigorous constraints in the early stages to refine the output and subsequently relaxing those constraints to protect critical information, the penalty ensures a stable convergence. This process allows the SCMA length to reach an equilibrium that minimizes redundancy while preserving essential logical integrity, thereby preventing the model from experiencing either excessive compression or numerical oscillation.

\section{Comparative Case Study of GRPO and SCMA}
As illustrated in the comparison between SCMA and GRPO presented in Case Study 1 and 2, our method demonstrates substantial advantages in reasoning efficiency and logical precision. By analyzing the red-highlighted redundant sections in the GRPO output, we can categorize the superiority of SCMA into the following dimensions:
\begin{itemize}
\item \textbf{Enhanced Logical Density and Convergence}: SCMA exhibits a high degree of logical density by maintaining a convergent reasoning path from the initial prompt to the final answer. While GRPO adopts a divergent thinking style that leads to unnecessary computational overhead, SCMA focuses on the essential deductive steps and avoids the generation of low-information tokens.
\item \textbf{Elimination of Redundant Meta-talk}: The response from GRPO is heavily populated with "meta-talk" such as filler phrases and transitional self-dialogue. Phrases like "Alright, let's break this down step by step" or "Wait, let me check that again" serve no functional purpose in solving the arithmetic problem. SCMA effectively prunes these linguistic redundancies and moves directly into the core mathematical logic.
\item \textbf{Suppression of Unproductive Circular Verification}: A critical weakness observed in the GRPO baseline is its tendency to engage in circular verification. As shown in the red-highlighted text, GRPO repeats the same calculation of "(16 - (3+4)) * 2" three separate times without producing any new insights. In contrast, SCMA performs a single, high-efficiency verification at the end of the process, which ensures accuracy while minimizing token consumption.
\item \textbf{Optimized Cognitive Processing of Simple Constraints}: GRPO demonstrates a form of "over-thinking" by questioning simple provided facts, such as the number of eggs used for muffins. This excessive skepticism leads to unproductive reasoning loops that increase latency. SCMA recognizes the straightforward nature of the constraints and processes them with appropriate cognitive intensity, thereby avoiding the analytical paralysis seen in the baseline.
\end{itemize}

\textbf{Summary of Improvements}: The primary distinction between the two models lies in the efficiency of their internal reasoning chains. In this specific case, approximately 40\% to 50\% of the tokens generated by GRPO are redundant or repetitive. SCMA succeeds in teaching the model that logical rigor does not require linguistic verbosity, which results in a reasoning process that is both faster and more cost-effective for real-world deployment.
% GRPO回答中冗余的部分主要体现在：反复确认题目条件、在已经得出结论后重新进行完整的计算流程、以及对题目意图进行过度的假设性怀疑。
\begin{modelcomparebox}[Case Study 1]

  % 第一部分：Prompt
Janet’s ducks lay 16 eggs per day. She eats three for breakfast every morning and bakes muffins for her friends every day with four. She sells the remainder at the farmers' market daily for \$2 per fresh duck egg. How much in dollars does she make every day at the farmers' market? Let's think step by step and output the final answer after "\#\#\#\#"

  % 第二部分：冗余回答
  \tcbsubtitle{Response of GRPO}
 \noindent \textless think\textgreater \\
Okay, let's see. So Janet's ducks lay 16 eggs each day. She uses some of them for breakfast and for baking muffins, and then sells the rest. The question is asking how much she makes every day at the farmers' market. \textcolor{red}{Alright, let's break this down step by step.}

\vspace{1em}

First, the total number of eggs laid per day is 16. Then, she eats three eggs for breakfast every morning. So that's 3 eggs taken out. Then, she bakes muffins for her friends every day with four eggs. \textcolor{red}{Wait, does that mean she uses four eggs for the muffins? So, total eggs used each day would be 3 for breakfast plus 4 for muffins. Let me check that again. The problem says she eats three for breakfast every morning and bakes muffins for her friends every day with four. So yeah, I think that's 3 + 4 = 7 eggs used each day.}

\vspace{1em}

So, total eggs laid: 16. Subtract the ones she uses: 16 - 7 = 9 eggs left to sell. Then, she sells the remainder at \$2 per egg. So, 9 eggs times \$2 per egg. Let me calculate that: 9 * 2 = \$18. So, she makes \$18 each day at the farmers' market.

\vspace{1em}

\textcolor{red}{Wait, let me double-check. Maybe I missed something. Let me go through each part again. Ducks lay 16 eggs per day. She eats 3 for breakfast. Then, she uses 4 eggs for muffins. So total eggs used are 3 + 4 = 7. 16 - 7 = 9. Then, 9 eggs sold at \$2 each. 9 * 2 is indeed 18. So, \$18 per day. That seems right.}

\vspace{1em}

\textcolor{red}{Is there any chance that the muffins require more eggs? Like, maybe four muffins per day? But the problem says she bakes muffins for her friends every day with four. So, I think that's four eggs per day for muffins. So, yeah, 3 + 4 = 7. So, 16 - 7 = 9. 9 * 2 = 18.}

\vspace{1em}

\textcolor{red}{I think that's correct. So the answer should be \$18.} \\
\textless /think\textgreater

  % 第三部分：凝练回答
  \tcbsubtitle{Response of SCMA}
  \noindent \textless think\textgreater \\
Okay, let's see. Janet's ducks lay 16 eggs each day. She uses some of them for breakfast and for muffins, and then sells the rest. I need to figure out how much she makes from selling the eggs.

\vspace{1em}

First, how many eggs does she use each day? She eats three for breakfast every morning. Then she bakes muffins with four eggs. So that's 3 plus 4, which is 7 eggs total that she uses.

\vspace{1em}

So if she starts with 16 eggs a day, subtract the 7 she uses, that leaves 16 minus 7. Let me calculate that: 16 - 7 equals 9 eggs. So she has 9 eggs left to sell each day.

\vspace{1em}

Now, each egg is sold for \$2. So 9 eggs times \$2 per egg. Let me do that multiplication: 9 * 2 = \$18.

\vspace{1em}

Therefore, she makes \$18 every day at the farmers' market. Let me just check my steps again to make sure I didn't make a mistake. Ducks lay 16 eggs. She uses 3 + 4 = 7. 16 - 7 = 9. 9 * 2 = 18. Yep, that seems right. I think that's the answer. \\
\textless /think\textgreater
\end{modelcomparebox}

\begin{modelcomparebox}[Case Study 2]

  % 第一部分：Prompt
A robe takes 2 bolts of blue fiber and half that much white fiber.  How many bolts in total does it take? Let's think step by step and output the final answer after "\#\#\#\#"

  % 第二部分：冗余回答
  \tcbsubtitle{Response of GRPO}
\noindent \texttt{<think>} \\
Okay, let's see. The problem says that a robe takes 2 bolts of blue fiber and half that much white fiber. I need to find out how many bolts in total it takes. \textcolor{red}{Hmm, let me break this down step by step.}

\vspace{1em}

\noindent First, the blue fiber part is straightforward. It takes 2 bolts. Now, the white fiber is half of that much. Wait, half of what? \textcolor{red}{The problem says ``half that much,'' so ``that'' probably refers to the 2 bolts of blue fiber.} So half of 2 bolts would be 1 bolt. So the white fiber requires 1 bolt.

\vspace{1em}

\noindent \textcolor{red}{Let me double-check that. If ``half that much'' refers to the blue fiber amount, which is 2 bolts, then yes, half of 2 is 1. So white fiber is 1 bolt.} Then the total bolts needed would be blue plus white, which is 2 + 1 = 3 bolts.

\vspace{1em}

\noindent \textcolor{red}{Wait, is there another way to interpret ``half that much''? Maybe someone might think ``that much'' refers to something else, but in the sentence structure, ``that'' is pointing back to the previous quantity mentioned, which is the 2 bolts of blue fiber. So I think my initial interpretation is correct.}

\vspace{1em}

\noindent \textcolor{red}{So adding them together: 2 bolts (blue) + 1 bolt (white) equals 3 bolts total. That seems right. Let me just make sure there's no trick or anything. The problem is pretty straightforward. No mention of any other fibers or additional requirements. So yeah, total bolts needed are 3.}

\vspace{1em}

\noindent \textcolor{red}{I think that's it. The answer should be 3.} \\
\texttt{</think>}

  % 第三部分：凝练回答
  \tcbsubtitle{Response of SCMA}
\noindent \texttt{<think>} \\
Okay, let's see. The problem says a robe takes 2 bolts of blue fiber and half that much white fiber. I need to find the total number of bolts required.
\vspace{1em}
\noindent First, the blue fiber is 2 bolts. Then the white fiber is half of that amount. Wait, half of what? Half of the blue fiber amount, right? So if the blue is 2 bolts, then half of that would be 1 bolt for the white fiber. So adding them together, 2 bolts for blue and 1 bolt for white makes a total of 3 bolts. Let me check that again. Half of 2 is 1, so yes, 2 + 1 equals 3. That seems right. I think the total is 3 bolts. \\
\texttt{</think>}
\end{modelcomparebox}

% \begin{promptbox}[title=System Prompt]
% You are a knowledgeable reasoning assistant. Your task is to decompose a given question into multiple sub-questions based purely on your own knowledge (do not use external tools), and answer each sub-question based on your own knowledge. If you do not know the answer, write 'unkown' exactly.\\

% % 注意：这里就像写普通文章一样，LaTeX 会自动处理第二行的对齐
% Output all results strictly in the following format for easy parsing:
% \textless{}q1\textgreater{}sub-question text\textless{}/q1\textgreater{}\\
% \textless{}a1\textgreater{}answer text\textless{}/a1\textgreater{}\\
% \textless{}q2\textgreater{}sub-question text\textless{}/q2\textgreater{}\\
% \textless{}a2\textgreater{}answer text\textless{}/a2\textgreater{}\\

% Continue numbering sequentially until done. At the end, output your final conclusion answer to the original question in the tag \textless{}predicted\_answer\textgreater{}...\textless{}/predicted\_answer\textgreater{}, and inside this tag follow the style of:
% [EXAMPLE\_PROMPT]
% \end{promptbox}
\end{document}